\documentclass[letterpaper]{article} 
\usepackage[]{aaai23}  
\usepackage{times}  
\usepackage{helvet}  
\usepackage{courier}  
\usepackage[hyphens]{url}  
\usepackage{graphicx} 
\usepackage{natbib}  
\usepackage{caption} 
\usepackage{algorithm}

\usepackage{newfloat}
\usepackage{listings}
\DeclareCaptionStyle{ruled}{labelfont=normalfont,labelsep=colon,strut=off} 
\floatstyle{ruled}
\newfloat{listing}{tb}{lst}{}
\floatname{listing}{Listing}
\usepackage{hyperref}       
\usepackage{url}            
\usepackage{booktabs}       
\usepackage{amsfonts}       
\usepackage{multirow}
\usepackage{lipsum}
\usepackage{indentfirst}
\usepackage{amsmath}
\usepackage{dsfont}
\usepackage{relsize}
\usepackage{svg}
\usepackage{amsthm}
\usepackage{xspace}
\usepackage{caption}
\usepackage{subcaption}
\usepackage{listings}
\usepackage{threeparttable}
\usepackage{array}
\usepackage{float}
\usepackage{ragged2e}
\usepackage{xcolor}
\usepackage{amssymb}
\usepackage{tabularx}
\usepackage{tikz}
\usepackage{algpseudocode}

\theoremstyle{definition}

\newcommand{\eg}{\textit{e.g.},\xspace}
\newcommand{\ie}{\textit{i.e.},\xspace}

\usepackage{lipsum}

\usepackage{listings}
\usepackage{xcolor}
\definecolor{codegreen}{rgb}{0,0.6,0}
\definecolor{codegray}{rgb}{0.5,0.5,0.5}
\definecolor{codepurple}{rgb}{0.58,0,0.82}
\definecolor{backcolour}{rgb}{0.99,0.99,0.98}
\lstdefinestyle{mystyle}{
    backgroundcolor=\color{backcolour},
    commentstyle=\color{codegreen},
    keywordstyle=\color{magenta},
    numberstyle=\tiny\color{codegray},
    stringstyle=\color{codepurple},
    basicstyle=\ttfamily\footnotesize,
    breakatwhitespace=false,
    breaklines=true,
    captionpos=b,
    keepspaces=true,
    numbers=left,
    numbersep=5pt,
    showspaces=false,
    showstringspaces=false,
    showtabs=false,
    tabsize=2,
    frame=shadowbox,
    rulesepcolor=\color{red!20!green!20!blue!20},
    xleftmargin=1em,xrightmargin=0em,aboveskip=1em,
    framexleftmargin=1em,
}

\title{Unveiling the Black Box of PLMs with Semantic Anchors:\\
Towards Interpretable Neural Semantic Parsing }
\author{
    Lunyiu Nie\textsuperscript{\rm 1}\equalcontrib \thanks{Work done during internship at Microsoft Research Asia.}, Jiuding Sun\textsuperscript{\rm 1}\equalcontrib,  Yanlin Wang\textsuperscript{\rm 2}\thanks{Yanlin Wang is the corresponding author. Work done during the author’s employment at Microsoft Research Asia.}, Lun Du\textsuperscript{\rm 3}, \\ Lei Hou\textsuperscript{\rm 1}, Juanzi Li\textsuperscript{\rm 1},  Shi Han\textsuperscript{\rm 3}, Dongmei Zhang\textsuperscript{\rm 3},  Jidong Zhai\textsuperscript{\rm 1} \\
}
\affiliations{
    \textsuperscript{\rm 1} Department of Computer Science and Technology, Tsinghua University\\

    \textsuperscript{\rm 2} School of Software Engineering, Sun Yat-sen University 
    \textsuperscript{\rm 3} Microsoft Research Asia \\ 
    \{nlx20, sjd22\}@mails.tsinghua.edu.cn, wangylin36@mail.sysu.edu.cn, \\ \{lun.du, shihan, dongmeiz\}@microsoft.com, \{houlei,lijuanzi, zhaijidong\}@tsinghua.edu.cn
}

\begin{document}

\maketitle

\begin{abstract}
The recent prevalence of pretrained language models (PLMs) has dramatically shifted the paradigm of semantic parsing, where the mapping from natural language utterances to structured logical forms is now formulated as a Seq2Seq task. Despite the promising performance, previous PLM-based approaches often suffer from hallucination problems due to their negligence of the structural information contained in the sentence, which essentially constitutes the key semantics of the logical forms. Furthermore, most works treat PLM as a black box in which the generation process of the target logical form is hidden beneath the decoder modules, which greatly hinders the model's intrinsic interpretability. To address these two issues, we propose to incorporate the current PLMs with a hierarchical decoder network. By taking the first-principle structures as the semantic anchors, we propose two novel intermediate supervision tasks, namely \textit{Semantic Anchor Extraction} and \textit{Semantic Anchor Alignment}, for training the hierarchical decoders and probing the model intermediate representations in a self-adaptive manner alongside the fine-tuning process. We conduct intensive experiments on several semantic parsing benchmarks and demonstrate that our approach can consistently outperform the baselines. More importantly, by analyzing the intermediate representations of the hierarchical decoders, our approach also makes a huge step toward the intrinsic interpretability of PLMs in the domain of semantic parsing. 
\end{abstract}

\section{Introduction}

\begin{figure}[t]
\centering
\resizebox{\columnwidth}{!}{
\includegraphics[]{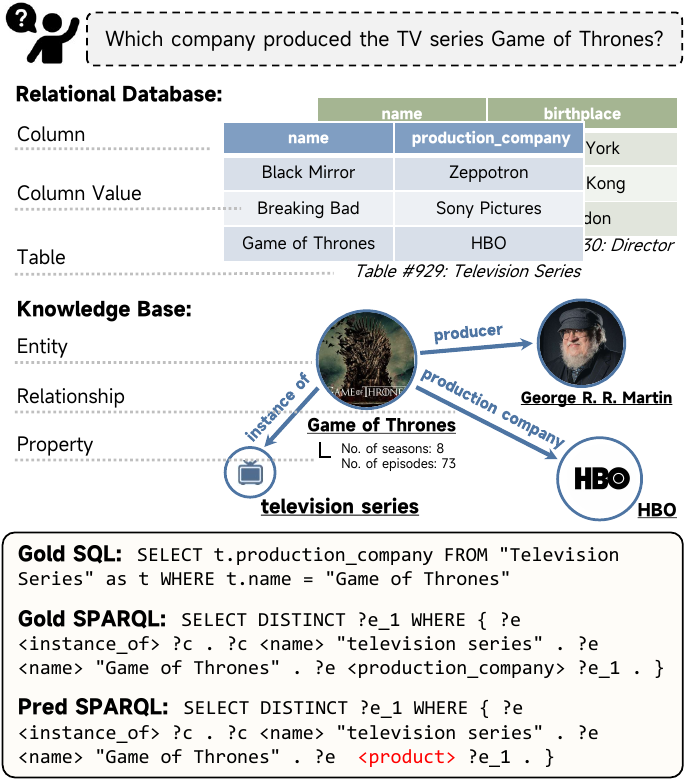}
}
\caption{Example of a natural language utterance and the corresponding SQL \& SPARQL logical forms. Specifically, the logical form sequences are composed of schema items that can be aligned to the structure of a database or a knowledge graph. Due to the negligence of these structures, PLM may suffer from hallucination issues and generate unfaithful information, as highlighted in the ``Pred SPARQL''. }
\label{fig:example}
\end{figure}

Semantic parsing refers to the task of converting natural language utterances into machine-executable logical forms \cite{kamath2018survey}. With the rise of pretrained language models (PLMs) in natural language processing, most recent works in the field formulate semantic parsing as a Seq2Seq task and develop neural semantic parsers on top of the latest PLMs like T5 \cite{raffel2020exploring}, BART \cite{lewis2020bart}, and GPT-3 \cite{brown2020language}, which significantly reduces the manual effort needed in designing compositional grammars \cite{liang2013learning, zettlemoyer2005learning}. By leveraging the extensive knowledge learned from the pretrain corpus, these PLM-based models exhibit strong performance in comprehending the semantics underlying the source natural language utterance and generating the target logical form that adheres to specific syntactic structures \cite{shin2021few, yin2021ingredients}.  

Despite the promising performance, current PLM-based approaches most regard both input and output as plain text sequences and neglect the structural information contained in the sentences \cite{yin2020tabert, shi2021learning}, such as the database (DB) or knowledge base (KB) schema that essentially constitutes the key semantics of the target SQL or SPARQL logical forms. As a result, these PLM-based models often suffer from the hallucination issue \cite{ji2022survey} and may generate incorrect logical form structures that are unfaithful to the input utterance \cite{nicosia2021translate, gupta2022retronlu}. For example, as shown in Figure \ref{fig:example}, the PLM mistakenly generates a relationship ``\texttt{product}'' in the SPARQL query, which is contradictory to the ``\textit{company produced}'' mentioned in the natural language.     

To prevent the PLMs from generating hallucinated structures, many works propose execution-guided decoding strategies \cite{wang2018robust, wang2021learning, ren2021lego} and grammar-constrained decoding algorithms \cite{shin2021constrained, scholak2021picard}. However, manipulating the decoding process with conditional branches can significantly slow down the model inference \cite{post2018fast, hui2021improving}. More importantly, in these methods, the DB/KB schema is employed extrinsically as a posteriori correction afterward the model fine-tuning, whereas the inherent ignorance of logical form structures still remains unsolved in the PLMs. 

Therefore, another concurrent line of work further pretrains the PLMs with structure-augmented objectives \cite{herzig2020tapas, deng2021structure}. Specifically, these works usually design unsupervised or weakly-supervised objectives for implicitly modeling the database structures with external or synthetic data corpus \cite{yu2020score, shi2022generation}. Although effective, further pretraining a large PLM can incur substantial costs and extra overheads \cite{yu2020grappa}. Besides, these methods also lack transferability since the structural knowledge is latently coupled inside the models and cannot be easily adapted to a novel task domain with a completely distinct database or knowledge base schema \cite{wu2021unified}. Thus, how to explicitly address the structural information during the PLM fine-tuning process is still an open question yet to be addressed. 

Aside from the above issue, existing neural semantic parsers typically treat PLMs as a black box lacking interpretability. Although some works attempt to probe and explain the latent knowledge within the PLMs using the external modules in a post hoc manner \cite{liu2021awakening, chen2021decoupled, stevens2021investigation}, none of the existing works explicitly addresses the intrinsic interpretability of neural semantic parsers. The intermediate process of logical form generation is completely hidden inside the PLM decoders, where the latent knowledge is hard to probe.

To address these challenges, we propose a novel model architecture with intermediate supervision over a hierarchical decoder network. Inspired by the first principle thinking and its successful application in AMR parsing \cite{cai2019core}, we define ``\textit{semantic anchors}'' as the building blocks of a logical form that cannot be further decomposed into more basic structures. For example, in a SQL query, semantic anchors include the tables (relations) and columns (attributes) that constitute the fundamental structure of a relational database \cite{aho1979theory, li2014constructing}; in a SPARQL query, semantic anchors include the entities, relationships, and their respective properties that similarly constitute the backbone of a knowledge base \cite{angles2008survey, barcelo2013querying}. 

Thereby, the semantic parsing process can now be broken down into the subtasks of extracting the semantic anchors from input utterances and subsequently recombining the identified semantic anchors into the target logical form based on certain formal syntax. We accordingly design two intermediate supervision tasks, namely \textit{Semantic Anchor Extraction} and \textit{Semantic Anchor Alignment}, for explicitly guiding the PLMs to address the structural information alongside the model fine-tuning process. Unlike the previous multi-task learning works that regard PLM as a whole \cite{radford2019language, aghajanyan2021muppet, xie2022unifiedskg}, we propose a hierarchical decoder architecture that self-adaptively attends to the PLM decoder layers for learning the intermediate supervision objectives. Eventually, this framework can equip the PLMs with intrinsic interpretability where the hidden representations of inner decoders originally concealed inside the PLMs are now unveiled for human analysis and investigation.  

Experimental results show that our proposed framework can consistently improve PLMs' performance on semantic parsing datasets \textsc{Overnight}, \textsc{KQA Pro} and \textsc{WikiSQL}. By investigating the inner representations of a PLM, our method also provides a novel testbed for interpreting the intermediate process of neural semantic parsing. In summary, our work contributes to the following aspects:

\begin{itemize}
    \item In this work, we summarize two major issues that hinder the neural semantic parsers: a) negligence of logical form structures, and b) lack of intrinsic interpretability. 
    \item To alleviate the problems, we propose a novel framework with hierarchical decoder and intermediate supervision tasks \textit{Semantic Anchor Extraction} and \textit{Semantic Anchor Alignment} that explicitly highlight the structural information alongside the PLM fine-tuning.
    \item By investigating the inner layer representations, this is also the first work in the field addressing the intrinsic interpretability of PLM-based semantic parsers. 
\end{itemize}
\begin{figure*}[t]
\centering
\resizebox{\textwidth}{!}{
\includegraphics[]{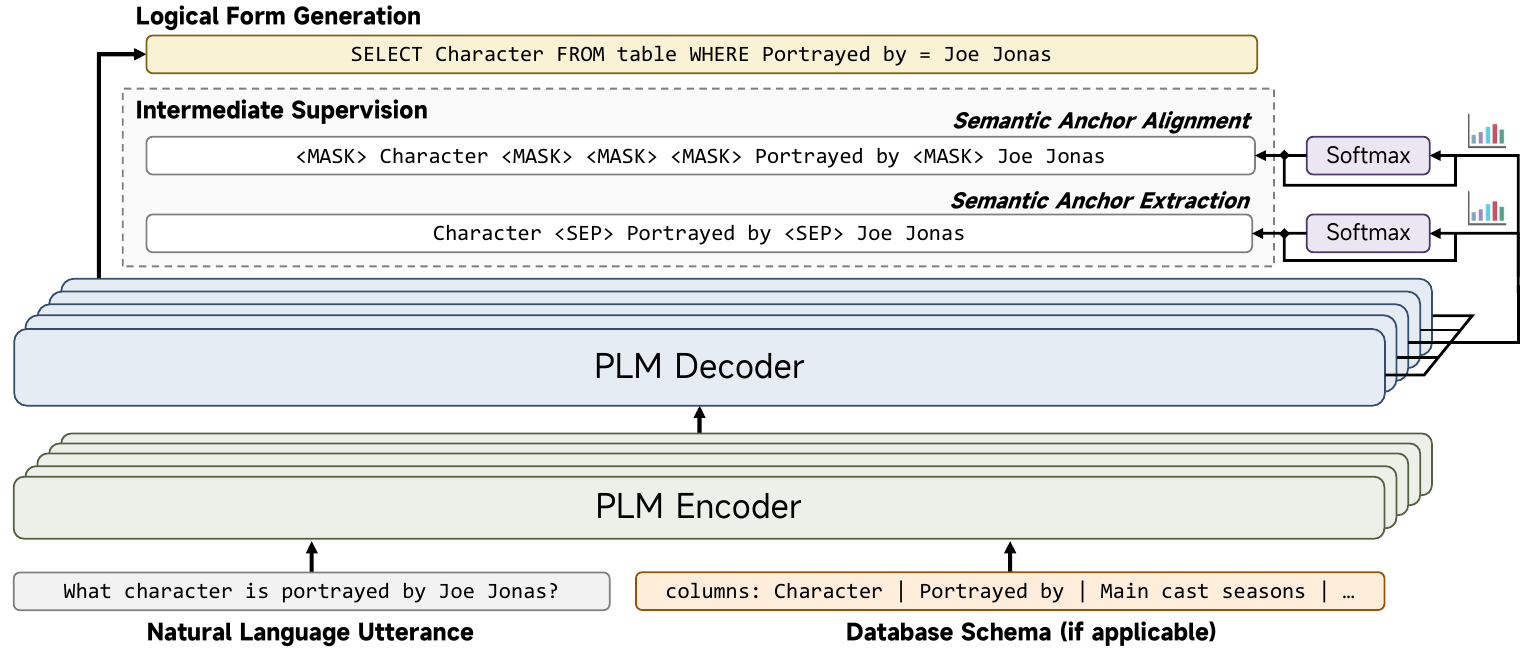}
}
\caption{Overall methodology of our proposed approach that incorporates two intermediate supervision tasks \textit{Semantic Anchor Extraction} and \textit{Semantic Anchor Alignment} into the intermediate PLM decoder layers for guiding \& probing the simultaneous main task fine-tuning of mapping the natural language utterance into a structured logical form. The middle part of the encoder and decoder modules can be flexibly replaced by any other PLMs. }
\label{fig:model}
\end{figure*}

\section{Methodology}
\subsection{Preliminaries}
In recent years, pretrained language models (PLMs) like BART \cite{lewis2020bart} and T5 \cite{raffel2020exploring} demonstrate strong generalization ability across various Seq2Seq tasks. Within these PLMs, the encoder module first projects the input sequence $\textbf{x}$ of length $m$ into a sequence of hidden states $\textbf{H}_{\mathbb{E}} = \{\textbf{h}_0, \textbf{h}_1, ..., \textbf{h}_m\}$ where each hidden state vector $\textbf{h}_{i}$ can be regarded as the contextual embedding of token $x_{i}$ in the high-dimensional space. 

Subsequently, the last encoder hidden states $\textbf{H}^{'}_{\mathbb{E}}$ is passed to the PLM decoder module consisting of $N$ layer of decoders. Each decoder layer simultaneously takes the previous decoder hidden states for self-attention computation and the last encoder hidden states for cross-attention computation \cite{vaswani2017attention} to produce the new hidden states:

\begin{equation}
    \textbf{H}_{\mathbb{D}}^{i} = \mathbf{Decoder}_{i}(\textbf{H}_{\mathbb{D}}^{i-1},\textbf{H}^{'}_{\mathbb{E}}\ |\ \theta^{i}_{\mathbb{D}}) 
\end{equation}

where $\theta^{i}_{\mathbb{D}}$ refers to the $i$-th decoder layer parameters and $\textbf{H}_{\mathbb{D}}^{i}$ is the corresponding output hidden states. Eventually, the last decoder hidden states $\textbf{H}_{\mathbb{D}}^{N}$ are projected into vocabulary-size $V$-dimensional logits by a linear layer and consequently generates the output tokens once at a time with greedy or beam search decoding. 

Therefore, neural semantic parsing can be formally defined as the mapping from a natural language sentence $\textbf{x}=\{x_1, x_2, ..., x_m\}$ into a logical form sequence $\textbf{y}=\{y_1, y_2, ..., y_k\}$ by maximizing the conditional probability over PLM parameters $\theta$:

\begin{equation}
    p(\textbf{y})=\prod_{i=1}^{k}p(y_i|\textbf{x}, y_1, y_2, ..., y_{i-1}; \theta).
\end{equation}.

\subsection{Semantic Anchor}
According to the above formulation, all tokens of a logical form sequence are treated equally by the PLMs, whereas the structural information inside the logical forms is neglected. To analyze the core structures contained in a logical form sequence, we start by giving the formal definitions of knowledge base and relational database.

\paragraph{Knowledge base}
A knowledge base (KB), often structured as an RDF graph or property graph, can be defined as a directed graph $G=(N, E)$ where $N$ is a set of nodes (or entities), $E$ is a set of edges (or relationships), $\lambda(N\cup E)\rightarrow L$ is a total function that defines the labels of all the nodes and edges, and $\sigma(N\cup E)\rightarrow (P,V)$ is a partial function that defines the (property, value) pairs of certain nodes or edges \cite{angles2017foundations}. 

Thereby, for any logical form $y$ querying a knowledge base, we formally define its \textbf{\textit{semantic anchors}} as the set of tokens corresponding to the knowledge base schema:
\begin{equation}
\mathcal{S}_{\textbf{y}|\text{KB}} = \{y_i \in \textbf{y}| y_i \subset (N\cup E\cup L\cup P\cup V)) \},
\end{equation}
including the KB entities, relationships, their respective labels, and applicable (property, value) pairs. 

\paragraph{Relational database}
A relational database (DB) is defined over a database schema $D$ including a set of relational schemas (or tables) $D=\{R_{i}|1\leq i\leq n\}$, where each relational schema further consists a set of attributes schemas (or columns) $R_{\textit{i}} = \{A_{j}^{i}|1\leq j\leq k\}$ \cite{li2014constructing}

Thereby, for any logical form $y$ querying a relational database, we formally define its \textbf{\textit{semantic anchors}} as the set of tokens aligned to the database schema: 

\begin{equation}
\mathcal{S}_{\textbf{y} | \text{DB}} = \{y_i \in \textbf{y}| y_i \subset (R \cup A)) \},
\end{equation}
including the DB table names and column names.

\subsection{Intermediate Supervision Tasks}
Based on the definition of \textit{semantic anchor}, we subsequently design two intermediate supervision tasks by decomposing the semantic parsing process into the subtasks of 1) extracting the \textit{semantic anchors} from the input natural language utterance, then 2) putting the extracted \textit{semantic anchors} into the right positions of a target sequence according to the syntax rule of target formal language. 

\paragraph{Semantic Anchor Extraction}
For the first intermediate supervision task, we enforce the PLMs to extract the \textit{semantic anchors} to explicitly address the logical form structures during the fine-tuning. For each logical form sequence $y$, we concatenate its \textit{semantic anchors} into a new sequence:  

\begin{equation}
    \textbf{y}_{\text{SAE}}=\{\mathcal{S}^{1}_{\textbf{y}}, \texttt{<SEP>}, \mathcal{S}^{2}_{\textbf{y}},\texttt{<SEP>}, ..., \mathcal{S}^{s}_{\textbf{y}}\}, 
\end{equation}
where \texttt{<SEP>} is a special token for separating two distinct \textit{semantic anchors}. A cross-entropy loss is calculated on this extraction supervision and the corresponding tokens at the inner decoder layers.


\paragraph{Semantic Anchor Alignment}
Thereafter as the second intermediate supervision task, we guide the model to generate the \textit{semantic anchors} with correct relative positions that can be precisely aligned to the final sequence of the target logical form. For each logical form sequence $y$, we only keep the \textit{semantic anchors} and mask the rest tokens:

\begin{equation}
    \textbf{y}_{\text{SAA}}=\{\texttt{<MASK>}, \texttt{<MASK>}, \mathcal{S}^{1}_{\textbf{y}}, \texttt{<MASK>}, ..., \mathcal{S}^{2}_{\textbf{y}}, ...\}
\end{equation}
where each \textit{semantic anchor} token $\mathcal{S}^{i}_{\textbf{y}}\in \mathcal{S}_{\textbf{y}}$ occurs in the exact relative position as aligned to the target logical form, and the remaining tokens masked by \texttt{<MASK>} are ignored during the loss computation.

\subsection{Hierarchical Decoders}
\label{sec: hierarchical}
To equip the PLMs with the ability to explicitly address the structural information and improve the intrinsic interpretability of the neural semantic parsers, we want to find a natural way to incorporate the \textit{semantic anchors} during the PLM fine-tuning. For a $N$-layers decoder module, the inner hidden states are given as $\{\textbf{H}_{\mathbb{D}}^{i}|1 \leq i \leq N-1\}$. 

For each intermediate supervision task $t$, we train an independent linear layer $f_{t}(\cdot)$ and a set of weighting parameters $\{w_{t}^{i}|1 \leq i \leq N-1\}$. Thereby,  we can calculate the aggregation of the inner decoder hidden states with a softmax distribution w.r.t. the weighting parameters and a residual connection:

\begin{equation}
    \textbf{H}_{\mathbb{D}}^{t} =  \underbrace{\sum_{i=1}^{N-1}\frac{e^{w_{t}^i}}{\sum_{j=1}^{N-1}e^{w_{t}^j}}\textbf{H}_{\mathbb{D}}^{i}}_{\text{learnable weights}} + \underbrace{\sum_{i=1}^{N-1}\frac{1}{N-1}\textbf{H}_{\mathbb{D}}^{i}}_{\text{residual connection}}
\end{equation}

This setup enables the model to attend to the inner decoder layers self-adaptively. The overall representation $\textbf{H}_{\mathbb{D}}^{s}$ is then mapped into logits in the vocabulary space $V$ by its task-respective linear layer:

\begin{equation}
    \textbf{\textit{v}}_t = f_{t}(\textbf{H}_{\mathbb{D}}^{t}), 
\end{equation}
then the probability token distribution can be given by a softmax function, and the cross-entropy loss $\mathcal{L}_{s}$ of the task can be computed consequently:

\begin{equation}
\begin{split}
    \mathcal{L}_{t} & = \sum \sum_{i=1}^{|v|}p(y_{t,i})\log p(\hat{y}_{i}) \\
    & = -\sum \sum_{i=1}^{|v|}y_{t,i}\log \frac{exp(\textbf{\textit{v}}_{t, i})}{\sum_{j=1}^{|v|}exp(\textbf{\textit{v}}_{t, j})}.
\end{split}
\end{equation}


\subsection{Self-adaptive Weighting}
Eventually, the overall fine-tuning of a PLM can now be defined as the aggregation of the main task (\ie logical form generation) and two intermediate supervision tasks:

\begin{equation}
    \mathcal{L}=\mathcal{L}_{main} + w_1 \mathcal{L}_{\text{SAE}} + w_2 \mathcal{L}_{\text{SAA}},
\end{equation}
where $w_{1}$ and $w_{2}$ denote the weighting factors for the two intermediate supervision tasks. To minimize the undesired interference between multiple learning objectives, we adopt a loss-balanced task weighting strategy to dynamically adjust the weighting factors throughout the fine-tuning process \cite{liu2019loss}. 

Specifically, for each intermediate supervision task $t$, we compute and store the first batch loss with respect to this task at each epoch, denoted as $\mathcal{L}_{(b_0, t)}$. The loss weighting factor $w_{t}$ is then dynamically adjusted at each iteration as:

\begin{equation}
    w_{t}=\sqrt{\frac{\mathcal{L}_{(b_j, t)}}{\mathcal{L}_{(b_0, t)}}},
\end{equation}
where $\mathcal{L}_{(b_j, t)}$ refers to the real-time loss of task $t$ at batch $j$.
\section{Experiments}
\subsection{Dataset}
\paragraph{Overnight} \textsc{Overnight} \cite{wang2015building} is a popular semantic parsing dataset containing 13,682 examples of natural language question paired with lambda-DCS logical forms across eight data domains so as to explore diverse types of language phenomena. We follow the previous practice \cite{cao2019semantic} and randomly sample 20\% of the provided training data as a validation set for performance evaluation during the PLM fine-tuning.

\paragraph{KQA Pro} \textsc{KQA Pro} \cite{shi2020kqa} is a KBQA dataset consisting of 117,790 natural language utterances and corresponded SPARQL queries over the Wikidata knowledge base \cite{vrandevcic2014wikidata}. It widely covers diverse natural language questions with explicitly enhanced linguistic variety and complex query patterns that involve multi-hop reasoning, value comparison, set operations, etc. 

\paragraph{WikiSQL} \textsc{WikiSQL} \cite{zhong2017seq2sql} is a classic Text-to-SQL semantic parsing dataset with 80,654 (question, SQL) data pairs grounded on 24,241 Wikipedia tables. Since WikiSQL queries cover only single tables and limited aggregators, previous PLM-based methods have almost achieved upper-bound performance on WikiSQL with the help of further pretraining and execution-guided decoding. Thus in this paper, we only compare to the models without using any additional resources or decoding-aiding techniques for fairness.

\subsection{Metric}
We use \textit{execution accuracy} as our evaluation metric. It examines whether the generated logical form can be executed by the respective KB or DB engines and return the exact set of results as identical to the ground truth logical forms.

\begin{table}[t!]
\centering
\small
\begin{tabular}{l c}
\toprule
  & \textbf{Exec. Acc.} \\ [0.5ex]
\midrule
\textbf{Non-PLM Methods} \\
    SPO \cite{wang2015building}     & 58.8 \\
    CrossDomain \cite{su2017cross}  & 80.6 \\
    Seq2Action \cite{chen2018sequence} & 79.0 \\
    2-stage DUAL  \cite{cao2020unsupervised} & 80.1 \\
    
    \midrule
    \textbf{PLM-based Methods} \\
    T5-base     & 74.7 \\
    Ours (T5-base)       & \textbf{75.5} \\
    \midrule
    BART-base                       & 80.7 \\
    GraphQ IR (BART-base) \cite{nie2022graphq}        & 82.1 \\
    Ours (BART-base)                         &  \textbf{82.4} \\
     \ w/o Semantic Anchor Extraction   & 81.0\\
     \ w/o Semantic Anchor Alignment    & 81.5\\
     \ w/o Hierarchical Decoder         & 81.2\\
     
\bottomrule
\end{tabular}
\caption{Test accuracies on \textsc{Overnight} dataset.}
\label{tab:overnight}
\end{table}

\begin{table}[t!]
\centering
\small
\begin{tabular}{l c}
\toprule
 & \textbf{Exec. Acc.} \\ [0.5ex]
\midrule
    \textbf{Non-PLM Methods} \\
    EmbedKGQA \cite{shi2020kqa}  & 28.36 \\
    RGCN \cite{shi2020kqa}       & 35.07  \\
    RNN \cite{shi2020kqa}        & 41.98 \\
    \midrule
    \textbf{PLM-based Methods} \\
    T5-base     & 83.64 \\
    Ours (T5-base)     & \textbf{84.66} \\
    \midrule
    BART-base \cite{shi2020kqa}   & 89.68 \\
    GraphQ IR (BART-base) \cite{nie2022graphq} & 91.70 \\
    Ours (BART-base)     & \textbf{91.72} \\
     \ w/o Semantic Anchor Extraction & 91.09\\
     \ w/o Semantic Anchor Alignment & 91.12\\
     \ w/o Hierarchical Decoder & 90.94\\
\bottomrule
\end{tabular}
\caption{Test accuracies on \textsc{KQA Pro} dataset.}
\label{tab:kqapro}
\end{table}

\begin{table}[h!]
\centering
\small
\begin{tabular}{l c }
\toprule
  & \textbf{Exec. Acc.} \\ [0.5ex]
\midrule
\textbf{Non-PLM Methods} \\
    Seq2SQL \cite{zhong2017seq2sql} & 59.4 \\
    Coarse-to-Fine \cite{dong2018coarse} & 78.5 \\
    Auxiliary Mapping \cite{chang2020zero} & 81.7 \\
    \midrule
    \textbf{PLM-based Methods} \\
    T5-base     & 84.5 \\
    Ours (T5-base)      & \textbf{85.0} \\
    \midrule
    BART-base   & 83.6 \\
    Ours (BART-base)    & \textbf{84.8} \\
     \ w/o Semantic Anchor Extraction & 84.3 \\
     \ w/o Semantic Anchor Alignment & 84.7 \\
     \ w/o Hierarchical Decoder & 84.2 \\
     \midrule
     \textbf{PLM Methods with Additional Resources} \\
     SQLova + EG \cite{hwang2019comprehensive} & 86.2 \\
     GRAPPA \cite{yu2020grappa} & 90.8 \\
     SeaD + EG \cite{xuan2021sead} & \textbf{93.0} \\
\bottomrule
\end{tabular}
\caption{Test execution accuracies on \textsc{WikiSQL} dataset. For fairness, we compare our methods with the plain-PLMs without using any additional resources (\eg further pretraining, data augmentation, execution-guided decoding, etc.). The SOTAs are also listed here for readers' information. }
\label{tab:wikisql}
\end{table}

\subsection{Experimental Settings}
We conduct our experiments with 8$\times$ NVIDIA Tesla V100 32GB GPUs and the CUDA environment of 10.2. All PLM models used in this work are acquired from the publicly released checkpoints on Huggingface \footnote{https://huggingface.co/models}. For BART-base, we fine-tuned the model with a learning rate of $3e-5$ and a warm-up proportion of 0.1. For T5-Base, the learning rate is set to $3e-4$ without warm-up. The batch size is consistently set to 128, and AdamW is used as the optimizer.

\subsection{Results}
Experiment results show that our proposed framework can consistently outperform the baselines on \textsc{Overnight}, \textsc{KQA Pro}, and \textsc{WikiSQL} datasets, as presented respectively in Table \ref{tab:overnight}, \ref{tab:kqapro}, and \ref{tab:wikisql}.  

Specifically, on both \textsc{Overnight} and \textsc{KQA Pro} datasets, our framework achieves the new state-of-the-art performance and demonstrates significant accuracy improvement over the PLM baselines. Remarkably, with the aid of intermediate supervision tasks and hierarchical decoder designs, our work even outperforms GraphQ IR \cite{nie2022graphq}, which requires the laborious implementation of an intermediate representation transpiler.  

On \textsc{WikiSQL}, our proposed approach also demonstrates superior execution accuracy over the T5-base and BART-base PLM baselines. In spite of the performance gap compared to the state-of-the-art works on \textsc{WikiSQL}, their works all rely heavily on further pertaining and execution-guided decoding, whereas our approach does not use any external resources other than the datasets themselves.

\begin{figure*}[h!]
\centering
\resizebox{\textwidth}{!}{
\includegraphics[]{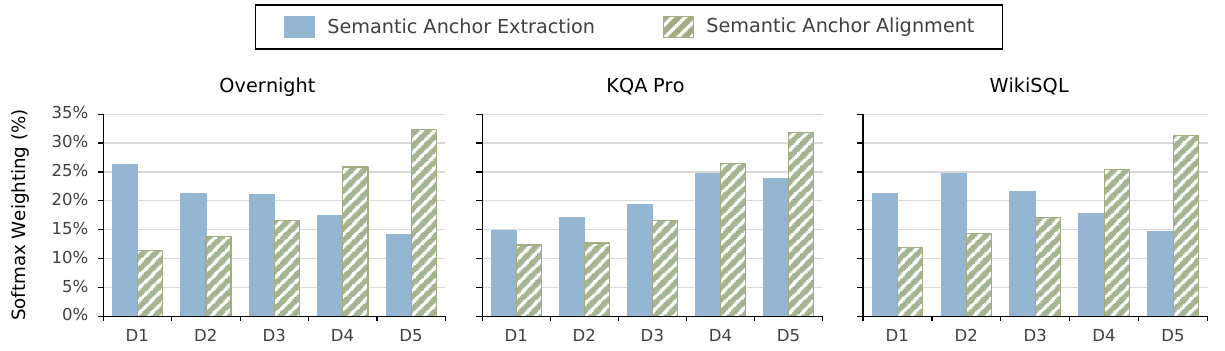}
}
\caption{Softmax distribution of the hierarchical decoders over the intermediate supervision tasks of \textit{Semantic Anchor Extraction} and \textit{Semantic Anchor Alignment}. D$_{i}$ refers to the weighting over the $i$-th intermediate decoder layer for the specified task.}
\label{fig:softmax}
\end{figure*}

\subsection{Ablation Studies}
For model ablation, we implement three different settings by removing the respective module: 
\begin{itemize}
    \item \textbf{Without Semantic Anchor Extraction} The proposed hierarchical decoder architecture with only the \textit{Semantic Anchor Alignment} task as the intermediate supervision. 
    \item \textbf{Without Semantic Anchor Alignment} The proposed hierarchical decoder architecture with only the \textit{Semantic Anchor Extraction} task as the intermediate supervision. 
    \item \textbf{Without Hierarchical Decoder} Both \textit{Semantic Anchor Extraction} and \textit{Semantic Anchor Alignment} tasks are performed at the top layer of the PLMs in a multi-task learning setting.
\end{itemize}

The ablated experiments demonstrate consistent trends across all of the benchmarks. Models without \textit{Semantic Anchor Extraction} demonstrate a larger drop in performance compared to those without \textit{Semantic Anchor Alignment}. This can be explained as similar to the human cognition process where the extraction of \textit{semantic anchors} is a more fundamental task as the premise of the latter constitution of a whole sequence. On the other hand, models without the hierarchical decoder also degrade significantly, which affirms our hypothesis that guiding the PLMs with supervision over the inner decoder layers can equip the models with improved robustness and intrinsic interpretability.

We notice that the \textit{Alignment} task poses larger impacts on KQA Pro (-0.9\%) and Overnight (-0.6\%) than on WikiSQL (-0.1\%). This can be explained by the relatively shorter logical form length in WikiSQL, which significantly ease the model's learning of the tokens’ position alignment.

\subsection{Hallucination Analysis}
\begin{figure}[h!]
\centering
\resizebox{\columnwidth}{!}{
\includegraphics[]{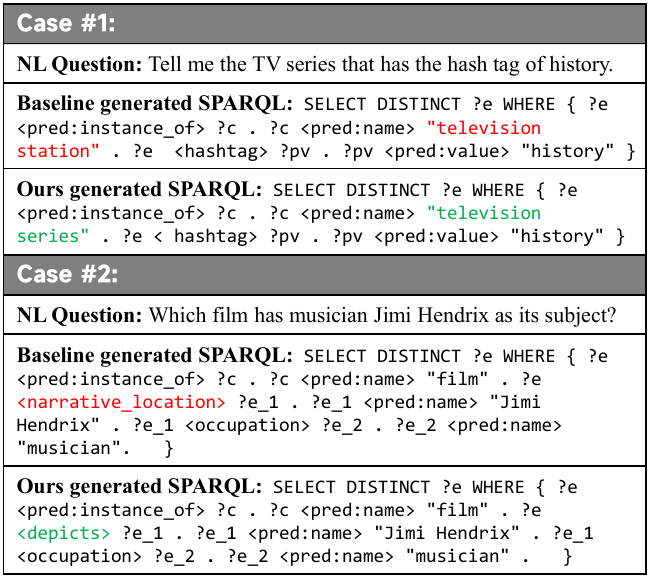}
}
\caption{Case analysis of the PLM hallucination issues. }
\label{fig:hallu}
\end{figure}

To call back the motivation and evaluate whether our proposed framework can help alleviate the hallucination issues in PLM-based semantic parsers, we further compare and analyze the generated logical forms from our method and from the BART-base baseline. We determine hallucination based on whether the generated logical form contains unfaithful or irrelevant schema tokens \cite{shi2021learning, ji2022survey}, and the results are shown in Table \ref{tab:hallu}. By explicitly addressing the \textit{semantic anchors} with intermediate supervision, our method can enforce the PLM to generate faithful structures and significantly reduce hallucination. Apart from the quantitative results, we also conduct case analysis and present two examples from the \textsc{KQA Pro} dataset in Figure \ref{fig:hallu}, where the PLM baseline mistakenly generates unfaithful content and our method can precisely output the correct SPARQL query.

\begin{table}[t]
\centering
\small
\begin{tabular}{l c c c}
\toprule
Dataset & Baseline & Ours & Difference\\
\midrule
\textsc{Overnight} & 294 & 278 & -5.76\% \\

\textsc{KQA Pro}  & 949 & 855 & -10.99\% \\

\textsc{WikiSQL} & 372 & 334 & -11.38\% \\

\bottomrule
\end{tabular}
\caption{Number of hallucination errors made by the BART-base baseline and our model.}
\label{tab:hallu}
\end{table}

\begin{figure*}[t]
\centering
\resizebox{\textwidth}{!}{
\includegraphics[]{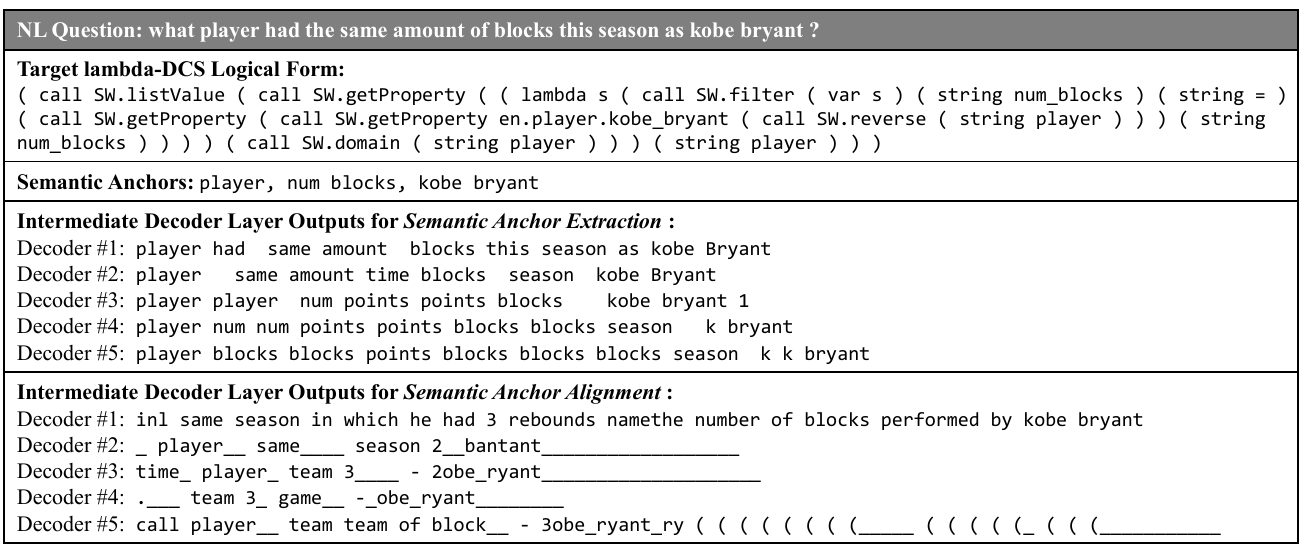}
}
\caption{BART-base inner decoder layer outputs respectively for the two intermediate supervision tasks. }
\label{fig:inter}
\end{figure*}

\subsection{Interpretability Analysis}
By performing intermediate supervision over the inner decoder layers together with the main task fine-tuning, this work also provides a novel testbed for probing the latent knowledge hidden inside a large PLM. Specifically, our framework can equip the PLMs with intrinsic interpretability in the following aspects.

\paragraph{Hierarchical Decoder Distribution} As aforementioned in Section \ref{sec: hierarchical}, during the intermediate supervision, the model can self-adaptively attend to the optimal inner decoder layers with weightings adjusted by loss backpropagation. Therefore, by analyzing the weighting distribution over the hierarchical decoders, we can thereby examine the sublayer functionalities of the PLM decoders. As can be observed in Figure \ref{fig:softmax}, PLMs tend to perform \textit{Extraction} at the lower decoder layers and \textit{Alignment} at the upper layers, which can be exactly aligned to the order of how humans may process the semantic parsing tasks. 

\paragraph{Intermediate Layer Output Analysis} More importantly, with our proposed hierarchical decoder, the PLM users are now able to probe the hidden representations of inner decoder layers. Specifically, by converting the layer-wise hidden logits into human-readable outputs, our work can be extended to understand the internal mechanisms behind PLM processing. We present an example from \textsc{Overnight} dataset with the BART-base inner 5 decoder layer outputs in Figure \ref{fig:inter}. We conclude that the lower layers of the PLM are more likely to contain information from the input sequence (\eg the Decoder \#1 outputs for both tasks are quite similar to the input natural language question). As the model hidden representations are further processed, the upper decoder layer outputs have moved closer to the target sequence (\eg the Decoder \#5 output for \textit{Semantic Anchor Alignment} already contain some syntax-related tokens like ``\texttt{call}'' and ``\texttt{(}'' in lambda-DCS logical forms). Some irrelevant tokens (\eg \texttt{points}) also exist in the inner layer output, which provides an indication for potential hallucination errors. Overall, these human-understandable inner layer outputs can greatly improve intrinsic interpretability by unveiling the latent representations from PLMs’ black box.
\section{Related Work}
\subsection{Semantic Parsing}
Semantic parsing is the task of converting natural language utterances into logical forms. 
The logical forms can be either downstream programs \cite{sun2020sparqa, zhong2017seq2sql,yu2018spider, shin2021few} or meaning representations \cite{banarescu2013abstract}.
Non-PLM-based methods in this field tend to tackle semantic parsing with representation learning or graph neural networks. \citeauthor{saxena2020improving} infuses knowledge representation to facilitate reasoning over the knowledge base.  \citeauthor{schlichtkrull2018rgcn} models the knowledge base as a graph by using GCN. Recent studies obtain state-of-the-art results on semantic parsing datasets with the capability of existing PLMs \cite{shin2021few, chen2021evaluating} or further pretrained domain-specific PLMs \cite{yin2020tabert, herzig2020tapas, yu2020grappa}. 



\subsection{PLM Interpretability}
Most of the studies attempt to reveal the latent connection in the PLM with post-hoc interpretation methods such as attention analysis \cite{clark2019does} and counterfactual manipulation \cite{stevens2020investigation}. \cite{yang2020sub} study the different functionalities of decoder sub-layers in transformer by analyzing the attention of each sub-layer while performing machine translation tasks. \cite{shi2020potential} take a more proactive approach to supervise the attention with prior knowledge during the training process and get a good result. \cite{liu2021awakening} explore the grounding capacity of PLMs by erasing the input tokens in sequential order to observe the change of confidence of each concept to be predicted.

\section{Conclusion}
In this paper, we address the two major issues inside the PLM-based semantic parsers. We design two intermediate supervision tasks, \textit{Semantic Anchor Extraction} and \textit{Semantic Anchor Alignment}, to guide the PLM training through a novel self-adaptive hierarchical decoder architecture. Extensive experiments show our framework can significantly reduce hallucination errors and demonstrate improved model intrinsic interpretability compared to the PLM baselines.


\newpage

\bibliography{aaai23}

\end{document}